%% file: main.tex
\documentclass[11pt,logo,onecolumn,copyright]{nvidiatechreport}
\usepackage[round]{natbib}
\usepackage{xspace}
\usepackage{placeins}
\usepackage{float}
\usepackage[most]{tcolorbox}
\usepackage{array}
\usepackage{ragged2e}
\usepackage{microtype}
\usepackage{multirow}
\usepackage{hyperref}
\definecolor{NvidiaGreen}{HTML}{76B900}
\definecolor{DraftGray}{HTML}{56616A}
\newcommand{\ourmodel}{Nemotron 3.5 Bolt E4B 4GB\xspace}
\newcommand{\ourmodelshort}{Bolt E4B\xspace}
\newcommand{\parentmodel}{Gemma-4-E4B-it\xspace}

\definecolor{TodoRed}{HTML}{B42318}
\newcommand{\nametodo}[1]
{\par\noindent{\small\color{TodoRed}\textbf{#1\if\relax\detokenize{#1}\relax\else: \fi TODO}}\par}

\setlist[itemize]{topsep=2pt,itemsep=2pt}
\renewcommand{\today}{2026-10-23}
\title{Nemotron 3.5 Bolt E4B 4GB:\\An Open and Efficient Small Language Model for Fast Conversational Agents}
\author{\large NVIDIA}
\date{}
\begin{document}

\begin{abstract}
\textbf{Abstract.} \input{sections/abstract}

\end{abstract}
\maketitle
\begin{center}
\end{center}
\input{sections/introduction}
\clearpage
\input{sections/chat_template}

\input{sections/post_training}

\input{sections/post_training_evaluation}

\input{sections/quantization}
\input{sections/quantization_evaluations}
\input{sections/multilingual}

\input{sections/multimodal}

\input{sections/speculative_decoding}
\input{sections/serving}
\input{sections/conclusion}
\input{sections/contributors}
\FloatBarrier
\bibliographystyle{references}
\bibliography{references}
\end{document}

%% file: sections/abstract.tex
We introduce \ourmodel, an open and efficient small language model further post-trained from Gemma-4-E4B-it and optimized for speed and intelligence per GB of memory used for fast conversational agents. The model carries forward Gemma's strong conversational capabilities and further improves multi-turn tool use, instruction following, and input grounding for fast, local voice/device assistants, interactive personas, and embedded/Edge interfaces. The model combines natural conversation with multi-turn agentic capabilities to power voice/device assistants, interactive characters and robotic/IoT interfaces running on GeForce RTX, Jetson, and RTX/DGX Spark.
Experiments on 21 benchmarks show that \ourmodel achieves the highest average accuracy while generating the fewest output tokens in both reasoning and non-reasoning modes, compared with \parentmodel and Qwen3.5-4B.
The model is released with a DFlash drafter.
With DFlash, \ourmodelshort achieves $2.74\times$ the
per-user decode speed of inference without speculative
decoding at concurrency~1, and $1.21\times$ the generation throughput of \parentmodel with MTP at batch size 1.

%% file: sections/introduction.tex
\section{Introduction}\label{sec:introduction}

\begin{figure}[ht]
\centering
\includegraphics[width=0.95\linewidth]{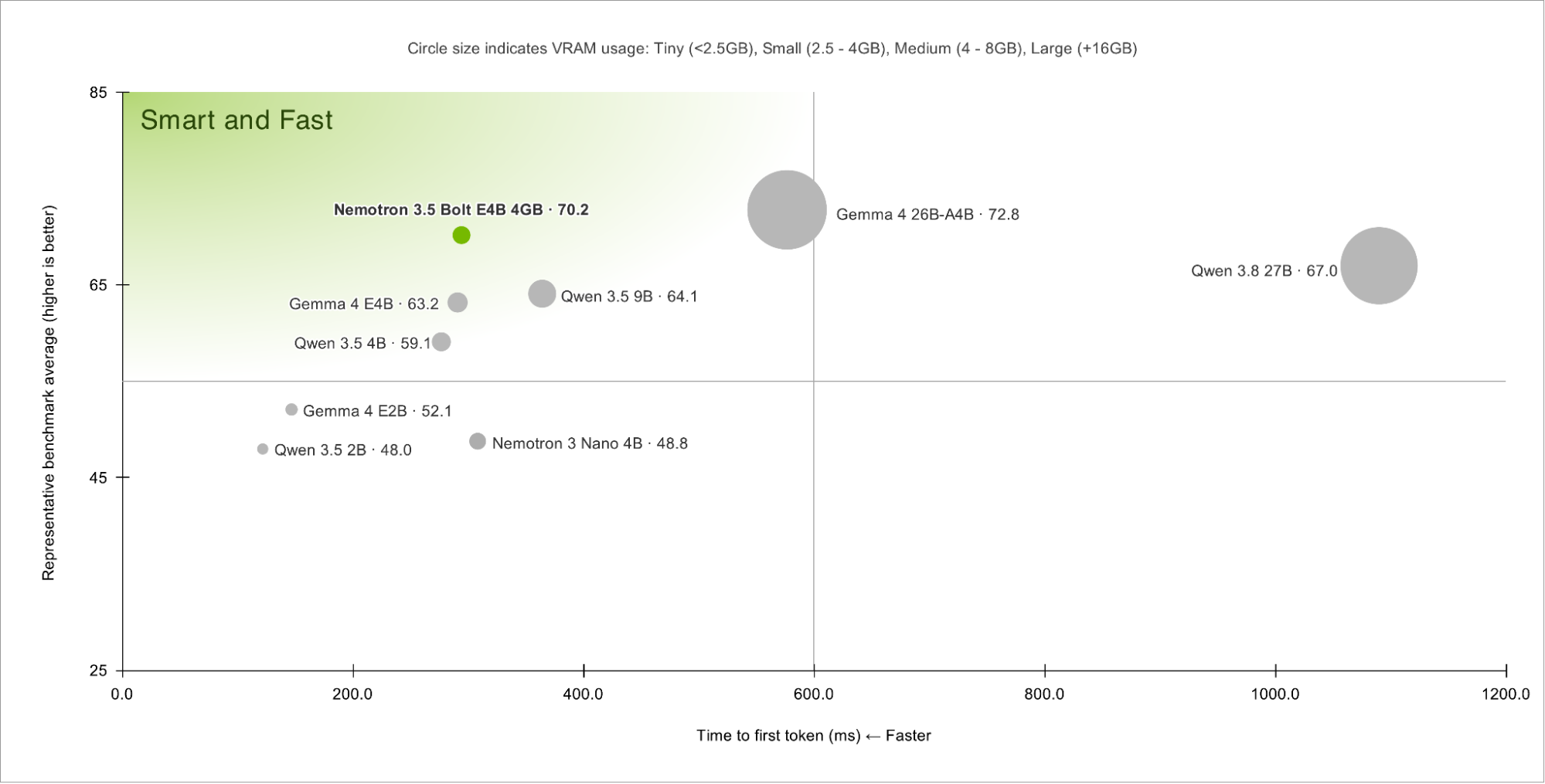}
\caption{Average benchmark score versus time to first token (TTFT). Scores are averaged across MultiChallenge, RoleMRC, BFCL-v3, IHEval, SysBench, and RelyToolBench. Local models use Q4\_K\_M quantization and are measured on RTX 5090 with input/output sequence lengths of approximately 3,000/50 tokens. Cloud reference results (score; TTFT): Gemini 3.8 Flash (77.1; 2,449.5ms) and Claude Sonnet 5.5 (76.2; 3,002.2ms).}
  \label{fig:accuracy-vs-ttft}
\end{figure}

Voice agents, interactive characters, and robotic interfaces need language models that can hold a conversation and take actions reliably. These models must follow instructions, use tools across multiple turns, and make sense of conversation history, application state, and tool results. For local deployment, they must also respond quickly and fit within limited memory.

We introduce \ourmodel, an open small language model designed for fast conversational agents with a target memory footprint of approximately 4 GB. \ourmodel{} supports reasoning and non-reasoning modes. The model can also generate concise conversational responses by enabling \textit{ conversational\_mode}. 

\ourmodel{} is based on Gemma4-E4B-it \citep{gemmateam2026gemma4} and further trained using RLVR and RLHF through \textsc{NeMo-RL} \citep{nemo-rl} and \textsc{NeMo-Gym} \citep{nemo-gym}. Training covers conversational tool use, tool-call schemas, instruction following, instruction hierarchy, and output formatting. We use every training sample in both reasoning and non-reasoning modes. This choice emphasizes accuracy without explicit reasoning while preserving reasoning capabilities for tasks that benefit from them. 

We evaluate \ourmodel on 21 benchmarks spanning conversational proficiency, multi-turn tool calling, instruction following/control, context robustness, and state/action grounding. Across this diverse evaluation suite, \ourmodel achieves the highest average benchmark score compared with Gemma4-E4B-it and Qwen3.5-4B in both reasoning and non-reasoning modes. It also generates fewer output tokens on average than either baseline in both reasoning and non-reasoning modes. These results demonstrate that \ourmodel is a strong fit for fast conversational agents.

For deployment, we quantize the BF16 checkpoint to Q4\_K\_M with importance-matrix calibration. We also provide a DFlash parallel drafter to accelerate generation through speculative decoding. The model is available via HF and NVIDIA NIM.

\paragraph{Checkpoints}
\begin{itemize}
    \item \href{https://huggingface.co/nvidia/NVIDIA-Nemotron-3.5-Bolt-E4B-4GB}{\texttt{NVIDIA-Nemotron-3.5-Bolt-E4B-4GB} \includegraphics[height=0.9em]{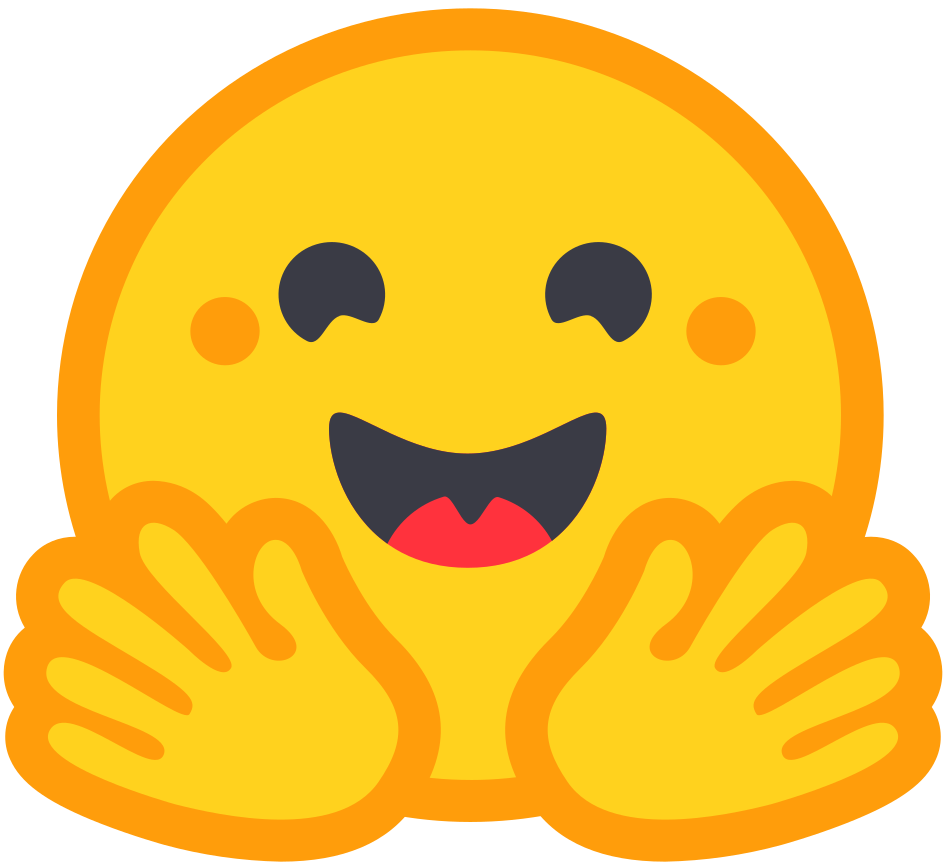}}:
    Q4\_K\_M-quantized model in GGUF format.

    \item \href{https://huggingface.co/nvidia/NVIDIA-Nemotron-3.5-Bolt-E4B-BF16}{\texttt{NVIDIA-Nemotron-3.5-Bolt-E4B-BF16} \includegraphics[height=0.9em]{figures/huggingface-color.png}}:
    BF16 model in Hugging Face format.

    \item \href{https://huggingface.co/nvidia/NVIDIA-Nemotron-3.5-Bolt-E4B-4GB-DFlash}{\texttt{NVIDIA-Nemotron-3.5-Bolt-E4B-4GB-DFlash} \includegraphics[height=0.9em]{figures/huggingface-color.png}}:
    DFlash drafter for the GGUF model, in GGUF format.

    \item \href{https://huggingface.co/nvidia/NVIDIA-Nemotron-3.5-Bolt-E4B-BF16-DFlash-NVFP4}{\texttt{NVIDIA-Nemotron-3.5-Bolt-E4B-BF16-DFlash-NVFP4} \includegraphics[height=0.9em]{figures/huggingface-color.png}}:
    DFlash drafter for the BF16 model, in HF format.
\end{itemize}

\paragraph{NVIDIA NIM}
\begin{itemize}
    \item \href{https://build.nvidia.com/nvidia/nemotron-3.5-bolt-e4b}{\texttt{build.nvidia.com/nvidia/nemotron-3.5-bolt-e4b}} : Start building with a free API endpoint.
    \item \href{https://catalog.ngc.nvidia.com/orgs/nim/nvidia/containers/nemotron-3.5-bolt-e4b-4gb/-}{\texttt{Nemotron-3.5-bolt-e4b-4gb}} : Download NVIDIA NIM for local deployment.
\end{itemize}

\paragraph{Data}
\begin{itemize}
    \item \href{https://huggingface.co/datasets/nvidia/Nemotron-System-User-Instruction-Hierarchy-v1}{\texttt{Nemotron-System-User-Instruction-Hierarchy-v1} \includegraphics[height=0.9em]{figures/huggingface-color.png}}:
    a collection of model-generated evaluation rubrics for assessing whether language-model responses follow the instruction hierarchy between system and user messages.
\end{itemize}

%% file: sections/chat_template.tex
\section{Chat Template Design}\label{sec:compression}

We extend the chat template of Gemma4-E4B-it to support alignment between tool-use and spoken responses in conversational applications.  When an assistant message contains text and tool calls, the template puts the text before the calls and their results. This order lets the assistant acknowledge a request or explain an action before it calls a tool. We keep the existing \textit{enable\_thinking} and \textit{preserve\_thinking} controls.

\subsection{Conversational Mode}

\ourmodel also supports an optional \textit{conversational\_mode} flag. When this flag is \textit{true}, the template adds a system instruction to keep responses concise and without excessive text formatting, making it more suitable for spoken conversations. The instruction is added after an existing system prompt, or if there is no system prompt, this instruction becomes the system prompt. When the \textit{conversational\_mode} flag is absent or \textit{false}, no instruction is added to the system prompt.

%% file: sections/post_training.tex
\section{Post-Training}

\ourmodel is further post-trained from Gemma4-E4B-it with multi-environment RLVR and RLHF. The collection of environments targets instruction following, multi-turn tool-calling and chat capabilities. The exact list of environments and link to the publicly released datasets is shown in Table \ref{tab:rl-envs}.

\begin{table}[h]
\caption{Environments and source datasets used in post-training for Nemotron-3.5-Bolt-E4B.}
\label{tab:rl-envs}
\centering
\scriptsize
\begin{tabular}{
  >{\raggedright\arraybackslash}p{0.2\textwidth}
  >{\raggedright\arraybackslash}p{0.75\textwidth}
}
\hline
\multicolumn{1}{l}{Environments} &
\multicolumn{1}{l}{Source data} \\
\hline

Tool-Calling &
\includegraphics[height=0.9em]{figures/huggingface-color.png} \href{https://huggingface.co/datasets/nvidia/Nemotron-RL-Agentic-Conversational-Tool-Use-Pivot-v1}{Conversational Tool Use Pivot} \\

Tool-call schema &
\includegraphics[height=0.9em]{figures/huggingface-color.png} \href{https://huggingface.co/datasets/nvidia/Nemotron-RL-Agentic-Function-Calling-Pivot-v1}{Function Calling Pivot} \\

Instruction Following &
\includegraphics[height=0.9em]{figures/huggingface-color.png} \href{https://huggingface.co/datasets/nvidia/Nemotron-RL-instruction_following}{Instruction Following}, \href{https://huggingface.co/datasets/nvidia/Nemotron-RL-SysBench-v1}{SysBench},
\href{https://huggingface.co/datasets/nvidia/Nemotron-RL-InverseIFEval-v1}{InverseIFEval},
\href{https://huggingface.co/datasets/nvidia/Nemotron-RL-CFBench-v1}{CFBench}, \href{https://huggingface.co/datasets/nvidia/Nemotron-RL-Instruction-Following-MultiTurnChat-v1}{MultiTurnChat}, \href{https://huggingface.co/datasets/nvidia/Nemotron-RL-Instruction-Following-Calendar-v2}{Calendar} \\

Instruction Hierarchy &
\includegraphics[height=0.9em]{figures/huggingface-color.png} \href{https://huggingface.co/datasets/nvidia/Nemotron-System-User-Instruction-Hierarchy-v1}{System-User-Instruction-Hierarchy} \\

Structured Outputs &
\includegraphics[height=0.9em]{figures/huggingface-color.png} \href{https://huggingface.co/datasets/nvidia/Nemotron-RL-Instruction-Following-Structured-Outputs-v2}{Structured Outputs} \\

Free-form formatting &
\includegraphics[height=0.9em]{figures/huggingface-color.png} \href{https://huggingface.co/datasets/nvidia/Nemotron-RL-Instruction-Following-Free-Form-Formatting-v1}{Free-Form Formatting} \\

GenRM & Updated version of \includegraphics[height=0.9em]{figures/huggingface-color.png} \href{https://huggingface.co/datasets/nvidia/Nemotron-RLHF-GenRM-v1}{Nemotron-RLHF-GenRM-v1} \\

\hline
\end{tabular}
\end{table}

The RLVR and RLHF post-training is done using \textsc{NeMo-RL} and \textsc{NeMo-Gym}. We use a global batch size of 256 with 8 rollouts per prompt, a maximum generation length of 4096 tokens for reasoning samples, and 1024 for non-reasoning samples.

\subsection{Non-Reasoning Focus}

To emphasize non-reasoning accuracy in post-training, we use each sample in our RLVR/RLHF training blend with both reasoning (\textit{enable\_thinking=true}) and non-reasoning (\textit{enable\_thinking=false}) inference modes. We independently perform reward profiling for the datasets in both reasoning modes, and maintain a deliberate 2 : 1 non-reasoning to reasoning samples ratio in each training batch.

\subsection{Verbosity Control}

We observe that as RL post-training progressed, there was a considerable increase in verbosity for both reasoning and final answer tokens. We were able to control verbosity by increasing group-relative length penalties for reasoning and answer length in the GenRM environment, as well as disabling overlong filtering for all other environments. While these interventions were effective in controlling verbosity, we found that tightening and loosening the length penalties, along with enabling and disabling overlong filtering in a cycle as the training progressed helped improve accuracy while still maintaining lower verbosity.

\subsection{SFT + RL vs. RL without SFT}

In an attempt to improve tool-calling behaviors of Gemma4-E4B-it, we experimented with SFT using datasets from \href{https://huggingface.co/collections/nvidia/nemotron-post-training-datasets}{Nemotron-Post-Training-Datasets}. We observed that we were able to improve multi-turn tool-calling with SFT and then further RLVR to a higher ceiling and with less RL training steps than with just RL-based post-training. However, SFT on data from diverse teachers resulted in a regression on conversational qualities and verbosity of Gemma4-E4B-it, for which it was used as a starting point. Hence, Nemotron-3.5-Bolt-E4B did not include an SFT stage and was post-trained purely with RL.

%% file: sections/post_training_evaluation.tex
\section{Post-Trained Model Evaluations}
\label{sec:evaluation}

\subsection{Benchmarks and Evaluation Setup}
We evaluate \ourmodel on a comprehensive suite of benchmarks covering conversational proficiency, multi-turn tool use, instruction following/control, context robustness, state/action grounding, and response conciseness. All evaluations for \ourmodel and baselines are conducted with \textsc{NeMo-Gym}. For each benchmark, we estimate the number of repeats needed to limit score variation to 1--2 percentage points and evaluate each model using that number of repeats. The following sections describe our benchmark selection and evaluation methodology across these capabilities, and Table \ref{tab:eval-config} presents the evaluation settings.

\begin{table}[!htbp]
\centering
\caption{Evaluation configuration for the benchmark suite. Reasoning mode indicates whether reasoning is enabled (On) or disabled (Off). Judge and User identify the evaluator and simulated user models, respectively.}\label{tab:eval-config}
\fontsize{8}{9.5}\selectfont
\setlength{\tabcolsep}{4pt}
\renewcommand{\arraystretch}{1.08}
\begin{tabularx}{\textwidth}{@{}>{\raggedright\arraybackslash}p{1.47in}|>{\raggedright\arraybackslash}p{1.60in}>{\centering\arraybackslash}p{0.57in}>{\raggedright\arraybackslash}X>{\centering\arraybackslash}p{0.38in}@{}}
\toprule
\textbf{Benchmark} & \textbf{Metric} & \textbf{Reason\-ing Mode} & \textbf{Judge/User Model} & \textbf{Re\-peats} \\
\midrule
\addlinespace[5pt]
\multicolumn{5}{@{}l}{\textbf{Conversational Proficiency}} \\
\midrule
MultiChallenge & Overall Score & On & Judge: GPT-4o & 16 \\
RoleBench Custom & Win \% vs. Gemma4-E4B-it & Off & Judge: GPT-5.5 & 4 \\
RoleMRC & Judge Average w/o Multi-Turn & On & Judge: GPT-4.1 & 8 \\
EQ-Bench v3 & Rubric Score & Off & Judge: Ensemble\textsuperscript{a} & 4 \\
Brevity Benchmark & Implicit - Average of Answer Quality \& Conciseness & On & User: Qwen3-235B-A22B-Instruct-2507 & 4 \\
\addlinespace[5pt]
\multicolumn{5}{@{}l}{\textbf{Multi-Turn Tool Use}} \\
\midrule
BFCL-v3 & Average Accuracy & On & N/A & 4 \\
When2Call Multi-Turn & Overall Accuracy & On & Judge: GPT-5.2 & 4 \\
ToolSandbox & Average Accuracy & On & User: GPT-4o & 4 \\
Tau2-Bench & Average Accuracy & On & User: GPT-5.2 & 4 \\
\addlinespace[5pt]
\multicolumn{5}{@{}l}{\textbf{Instruction Following \& Control}} \\
\midrule
IHEval & Conflict Accuracy & On & N/A & 2 \\
AgentIF & Instruction Success Rate (ISR) & On & Judge: GPT-5.1 & 4 \\
StructEval-T & Average Accuracy & On & N/A & 4 \\
IFBench & Prompt Loose Accuracy & On & N/A & 4 \\
InverseIFEval & Overall Accuracy & On & Judge: GPT-5.1 & 8 \\
\addlinespace[5pt]
\multicolumn{5}{@{}l}{\textbf{Context Robustness}} \\
\midrule
RULER (64K) & Overall Accuracy & Off & N/A & 4 \\
LongMemEval & Overall Accuracy & On & Judge: GPT-4o & 8 \\
LongBench v2 (48K) & Overall Accuracy & On & N/A & 4 \\
SysBench & Instruction Success Rate (ISR) & On & Judge: GPT-5.1 & 4 \\
\addlinespace[5pt]
\multicolumn{5}{@{}l}{\textbf{State \& Action Grounding}} \\
\midrule
GaRAGe & Relevance-Aware Factuality (RAF) & On & Judge: GPT-5.1 & 4 \\
RAGTruth & F1 & On & N/A & 4 \\
RelyToolBench\newline w/ Native Function-Calling & Real Pass Rate (RePR) & On & Judge: GPT-5.1 & 8 \\
\bottomrule
\end{tabularx}
\par\smallskip
\begin{minipage}{\textwidth}
\fontsize{8}{9.5}\selectfont
\textsuperscript{a}Ensemble: Claude-Sonnet-5.5, Gemini-4-Argon, and GPT-6.1-Sol.
\end{minipage}
\end{table}

\begin{table}[!htbp]
\centering
 \caption{Evaluation results for BF16 checkpoints. Evaluation metrics and reasoning settings are listed in Table~\ref{tab:eval-config}. Bold marks the best score in each row.}\label{tab:bf16-evals}
\fontsize{9}{10.7}\selectfont
\setlength{\tabcolsep}{5pt}
\renewcommand{\arraystretch}{1.08}
\begin{tabularx}{\textwidth}{@{}>{\raggedright\arraybackslash}X|rrrr@{}}
\toprule
\textbf{Benchmark} &
\shortstack{\textbf{Nemotron-3.5}\\\textbf{Bolt-E4B}} &
\shortstack{\textbf{Gemma4}\\\textbf{E4B-it}} &
\shortstack{\textbf{Qwen3.5}\\\textbf{4B}} &
\shortstack{\textbf{Nemotron-3}\\\textbf{Nano-4B}} \\
\midrule
\addlinespace[5pt]
\multicolumn{5}{@{}l}{\textbf{Conversational Proficiency}} \\
\midrule
MultiChallenge & \textbf{53.10} & 42.08 & 51.76 & 35.30 \\
RoleBench Custom & \textbf{67.05} & 49.95 & 53.66 & 12.02 \\
RoleMRC & \textbf{80.63} & 74.62 & 65.81 & 47.00 \\
EQ-Bench v3 & 39.40 & \textbf{41.17} & 32.92 & 19.31 \\
Brevity Benchmark & \textbf{82.52} & 79.99 & 75.35 & 73.60 \\
\addlinespace[5pt]
\multicolumn{5}{@{}l}{\textbf{Multi-Turn Tool Use}} \\
\midrule
BFCL-v3 & 68.51 & 66.39 & \textbf{70.95} & 61.10 \\
When2Call Multi-Turn & \textbf{63.36} & 62.90 & 37.85 & 27.70 \\
ToolSandbox & 68.77 & 65.43 & \textbf{69.54} & 57.88 \\
Tau2-Bench & 55.10 & 38.50 & \textbf{75.06} & 29.20 \\
\addlinespace[5pt]
\multicolumn{5}{@{}l}{\textbf{Instruction Following \& Control}} \\
\midrule
IHEval & \textbf{84.79} & 67.04 & 60.57 & 51.31 \\
AgentIF & \textbf{25.81} & 21.89 & 20.44 & 20.08 \\
StructEval-T & \textbf{80.40} & 79.27 & 76.01 & 63.50 \\
IFBench & \textbf{63.79} & 42.84 & 58.91 & 60.10 \\
InverseIFEval & \textbf{46.84} & 42.00 & 30.98 & 35.25 \\
\addlinespace[5pt]
\multicolumn{5}{@{}l}{\textbf{Context Robustness}} \\
\midrule
RULER (64K) & 86.99 & 86.55 & \textbf{95.15} & 91.51 \\
LongMemEval & \textbf{72.10} & 70.95 & 66.85 & 25.20 \\
LongBench v2 (48K) & 50.50 & 49.34 & \textbf{59.93} & 31.40 \\
SysBench & \textbf{67.88} & 65.85 & 67.34 & 60.88 \\
\addlinespace[5pt]
\multicolumn{5}{@{}l}{\textbf{State \& Action Grounding}} \\
\midrule
GaRAGe & \textbf{42.61} & 34.73 & 24.66 & 25.18 \\
RAGTruth & \textbf{77.59} & 76.97 & 75.60 & 71.40 \\
RelyToolBench & \textbf{63.16} & 61.60 & 43.35 & 41.20 \\
\bottomrule
\end{tabularx}
\end{table}

\paragraph{Conversational Proficiency.}
We evaluate conversational proficiency using {\bf EQ-Bench v3} \citep{eqbench3,eqbench} for emotional understanding, empathy, and social reasoning, with an ensemble of Claude-Sonnet-5.5, Gemini-4-Argon, and GPT-6.1-Sol judges; {\bf MultiChallenge} \citep{multichallenge} for instruction retention, context tracking, and response consistency across multiple turns, judged by GPT-4o; {\bf RoleBench} \citep{rolellm} for role-playing fidelity and character-consistent responses, judged by GPT-5.5; and {\bf RoleMRC} \citep{rolemrc} for role-conditioned reading comprehension and instruction following, judged by GPT-4.1. For RoleBench, we use a custom pairwise evaluation against a fixed Gemma4-E4B-it reference, with each response pair judged in both answer orders to reduce position bias. For RoleMRC, we report the average judge score excluding the multi-turn subset.

\paragraph{Response Conciseness.}
As part of conversational proficiency, we also evaluate response conciseness using an internal benchmark for brevity which covers personal assistant, customer service, and passage-based question-answering tasks. The benchmark compares response length, excluding reasoning tokens, with human-written reference responses, awarding high scores to responses that are no longer than the references. System prompts request concise answers through general instructions such as ``brief'' or ``suitable for spoken conversation'', making conciseness an instruction-following task. Answer quality is assessed separately by an LLM judge to check that shorter responses preserve quality. We report the average of answer quality and conciseness scores. The benchmark will be publicly released in the future. We also use the results from this benchmark to determine the system prompt to be used for \textit{conversational\_mode} described in the chat template design section.

\paragraph{Multi-Turn Tool Use.}
We evaluate multi-turn tool use using {\bf BFCL-v3} \citep{bfcl,bfclv3} for fundamental tool calling; {\bf $\tau^2$-Bench} \citep{tau2bench} for task completion through tool use and dialogue, with GPT-5.2 as the user model; and {\bf ToolSandbox} \citep{toolsandbox} for stateful tool use in conversational settings, with GPT-4o as the user model. We also evaluate {\bf When2Call-MT}, our multi-turn extension of When2Call \citep{when2call}, which tests whether an assistant selects the correct response policy as a conversation develops, including when to call a tool, ask for clarification, refuse due to lack of tools, or respond directly. When2Call-MT uses GPT-5.2 as the judge and will be publicly released in the future.

\paragraph{Instruction Following \& Control.}
We evaluate instruction following and control using {\bf AgentIF} \citep{agentif} for following complex instructions in agentic settings, judged by GPT-5.1; {\bf IFBench} \citep{ifbench} for generalization to verifiable instruction-following constraints; {\bf IHEval} \citep{iheval} for resolving conflicting instructions according to the instruction hierarchy; InverseIFEval \citep{inverseifeval} for following counter-intuitive instructions, judged by GPT-5.1; and StructEval-T \citep{structeval} for generating outputs that satisfy specified structural and formatting requirements. For {\bf InverseIFEval}, we evaluate using the methodology as originally described, but use only English data (506 of 1012 total samples) and translate the judge prompt to English.

\paragraph{Context Robustness.}
We evaluate context robustness using {\bf LongBench v2} \citep{longbenchv2} for understanding and reasoning over long contexts, restricted to the subset with fewer than 48K tokens due to the model's context limit; {\bf LongMemEval} \citep{longmemeval} for retaining and retrieving information across conversation histories, judged by GPT-4o; {\bf RULER} \citep{ruler} for retrieval and reasoning at a 64K-token context length; and {\bf SysBench} \citep{sysbench} for following system instructions across varied user inputs, judged by GPT-5.1. For SysBench, we translate the benchmark data into English and use an adapted English judge prompt.

\paragraph{State \& Action Grounding.}
We evaluate state and action grounding using {\bf GaRAGe} \citep{garage} for retrieved evidence-grounded answers, judged by GPT-5.1; {\bf RAGTruth} \citep{ragtruth} for detecting unsupported claims in retrieval-augmented responses, measured by span-level F1; and {\bf RelyToolBench} \citep{relytoolbench} for reliable tool use under tool failures and misleading tool responses. We evaluate RelyToolBench with native function calling instead of the ReAct format and use GPT-5.1 as the judge. For GaRAGe, we follow the evaluation methodology as originally described and use our own implementation due to a lack of publicly available code.

\subsection{Results}
We show all benchmark evaluations in Table \ref{tab:bf16-evals}.
Our model outperforms the competitive models on 15 out of 21 benchmarks: Conversational Proficiency (4/5), Multi-Turn Tool Use (1/4), Instruction Following \& Control (5/5), Context Robustness (2/4), and State \& Action Grounding (3/3).

\paragraph{Non-Reasoning Evaluations.}
We also conduct all benchmark evaluations in non-reasoning mode across models to validate the effect of focused post-training. The results are presented in Table \ref{tab:non-reasoning-bf16-evals}. The results show a similar trend as reasoning, and \ourmodel shows the highest accuracy on 14 out of 21 benchmarks.

\begin{table}[!htbp]
\centering
\caption{Non-reasoning evaluation results for BF16 checkpoints. Bold marks the best score in each row, including ties.}\label{tab:non-reasoning-bf16-evals}
\fontsize{9}{10.7}\selectfont
\setlength{\tabcolsep}{5pt}
\renewcommand{\arraystretch}{1.08}
\begin{tabularx}{\textwidth}{@{}>{\raggedright\arraybackslash}X|cccc@{}}
\toprule
\textbf{Benchmark} &
\shortstack{\textbf{Nemotron-3.5}\\\textbf{Bolt-E4B}} &
\shortstack{\textbf{Gemma4}\\\textbf{E4B-it}} &
\shortstack{\textbf{Qwen3.5}\\\textbf{4B}} &
\shortstack{\textbf{Nemotron-3}\\\textbf{Nano-4B}} \\
\midrule
\addlinespace[5pt]
\multicolumn{5}{@{}l}{\textbf{Conversational Proficiency}} \\
\midrule
MultiChallenge & \textbf{48.81} & 40.72 & 38.60 & 23.27 \\
RoleBench Custom & \textbf{67.05} & 49.95 & 53.66 & 12.02 \\
RoleMRC & \textbf{77.18} & 72.46 & 58.91 & 46.11 \\
EQ-Bench v3 & 39.40 & \textbf{41.17} & 32.92 & 19.31 \\
Brevity Benchmark & \textbf{78.71} & 75.39 & 75.21 & 72.43 \\
\addlinespace[5pt]
\multicolumn{5}{@{}l}{\textbf{Multi-Turn Tool Use}} \\
\midrule
BFCL-v3 & 64.68 & 58.08 & \textbf{66.04} & 45.23 \\
When2Call Multi-Turn & 51.25 & \textbf{54.80} & 46.02 & 38.67 \\
ToolSandbox & \textbf{61.54} & 35.72 & 52.19 & 55.32 \\
Tau2-Bench & 34.34 & 25.99 & \textbf{73.46} & 26.78 \\
\addlinespace[5pt]
\multicolumn{5}{@{}l}{\textbf{Instruction Following \& Control}} \\
\midrule
IHEval & \textbf{66.90} & 52.95 & 34.25 & 23.78 \\
AgentIF & \textbf{23.76} & 22.98 & 21.22 & 16.69 \\
StructEval-T & \textbf{78.76} & 76.04 & 75.84 & 10.38 \\
IFBench & \textbf{53.33} & 38.83 & 34.33 & 41.25 \\
InverseIFEval & \textbf{30.24} & 22.13 & 22.94 & 18.97 \\
\addlinespace[5pt]
\multicolumn{5}{@{}l}{\textbf{Context Robustness}} \\
\midrule
RULER (64K) & 86.99 & 86.55 & \textbf{95.15} & 91.51 \\
LongMemEval & 60.20 & \textbf{65.79} & 65.00 & 31.20 \\
LongBench v2 (48K) & 41.06 & 41.89 & \textbf{51.70} & 31.13 \\
SysBench & \textbf{65.12} & 64.09 & \textbf{65.12} & 55.28 \\
\addlinespace[5pt]
\multicolumn{5}{@{}l}{\textbf{State \& Action Grounding}} \\
\midrule
GaRAGe & \textbf{46.04} & 37.33 & 22.19 & 23.61 \\
RAGTruth & \textbf{67.38} & 58.20 & 65.08 & 30.87 \\
RelyToolBench & \textbf{61.25} & 58.87 & 19.43 & 36.18 \\
\bottomrule
\end{tabularx}
\end{table}

\begin{figure}[t]
  \centering
  \includegraphics[width=0.95\linewidth]{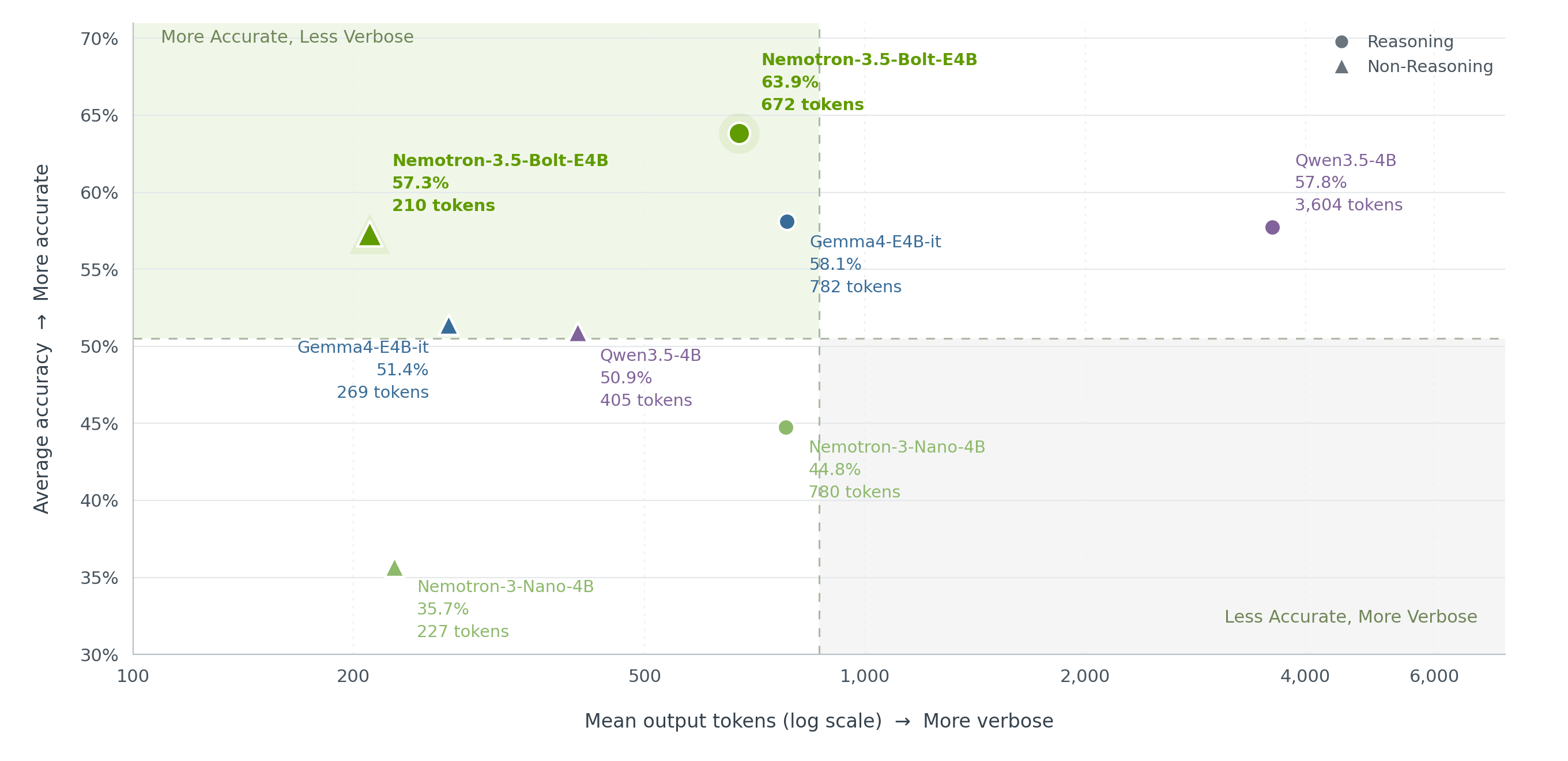}
  \caption{Average benchmark score versus mean output token count across 21 benchmarks in reasoning and non-reasoning settings. Higher scores and fewer output tokens indicate higher quality and greater response conciseness, respectively. The horizontal axis uses a logarithmic scale.}
  \label{fig:accuracy-vs-output-tokens}
\end{figure}

\subsection{Live Testing}
\begin{figure}[t]
\centering
\begin{minipage}[t]{0.48\textwidth}
    \centering
    \includegraphics[height=4.5cm,width=\linewidth]{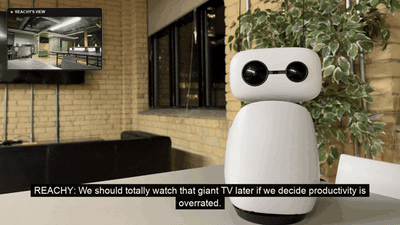}

    \par\vspace{3pt}
    {\justifying\footnotesize Reachy Mini running on Jetson Orin Nano 8GB. Cosmos 3 Edge \citep{nvidia2026cosmos3} supplies real-time visual understanding of the world to Bolt E4B, which calls scripts from the robot's animation libraries and powers its spoken responses.\par}
\end{minipage}
\hfill
\begin{minipage}[t]{0.48\textwidth}
    \centering
    \includegraphics[height=4.5cm,width=\linewidth]{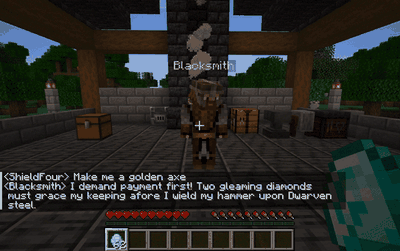}

    \par\vspace{3pt}
    {\justifying\footnotesize Bolt E4B controlling a blacksmith trader NPC. The NPC agent verifies correct payment for an item, then autonomously 
calls tools to check crafting recipes, collect ingredients, craft the item, and transfer it to the user.\par}
\end{minipage}
\caption{Examples of live-testing conducted with \ourmodel. View the GIFs in our \href{https://research.nvidia.com/labs/nemotron/NVIDIA-Nemotron-3.5-Bolt-E4B-4GB/}{blog.}}
\label{fig:live-testing-examples}
\end{figure}

Academic benchmarks only partially capture the skills required of conversational agents and typically assess individual capabilities rather than how they work together over a complete interaction. We therefore complement benchmark evaluations with live testing in interactive scenarios spanning virtual interactive characters, on-device voice assistants, and chat interactions. These tests are qualitative and examine whether the model can sustain coherent conversations, respond to changing user intent, environment feedback, handle ambiguity, and recover from mistakes or tool failures. GIFs in Figure \ref{fig:live-testing-examples} shows captures from some live-testing instances for \ourmodel.

%% file: sections/quantization.tex
\section{Quantization}
We used llama.cpp to convert the \ourmodel BF16 checkpoint to Q4\_K\_M with importance-matrix (imatrix) calibration. The calibration was conducted with approximately 1M tokens of RL rollouts from the training step that was selected as the BF16 checkpoint. On average accuracy, Q4\_K\_M with imatrix calibration not only outperforms vanilla Q4\_K\_M, but also IQ4\_XS and IQ4\_NL with imatrix calibration. Evaluation results for quantized models are shown in Table \ref{tab:q4-evals}.

%% file: sections/quantization_evaluations.tex
\providecommand{\singleheader}[1]{\raisebox{0.6ex}{#1}}
\begin{table*}[!ht]
\caption{Evaluation results for four-bit quantized checkpoints. The left block compares Q4\_K\_M checkpoints, with importance-matrix calibration (imatrix) used for Bolt. We use the Unsloth Q4\_K\_M GGUF checkpoints for
Gemma4-E4B-it and Qwen3.5-4B.\protect\footnotemark. The right block shows additional Bolt variants: Q4\_K\_M without imatrix, IQ4\_XS, and IQ4\_NL. All benchmarks use reasoning mode except EQ-Bench v3, RoleBench Custom, and RULER (64K), which use non-reasoning mode.}
\centering
\small
\setlength{\tabcolsep}{3pt}
\fontsize{8}{9.5}\selectfont
\renewcommand{\arraystretch}{1.08}
\begin{tabular}{l|cccc|ccc}
\hline
 & \multicolumn{4}{c|}{\textbf{Q4\_K\_M model comparison}}
 & \multicolumn{3}{c}{\textbf{Nemotron-3.5-Bolt-E4B Q4 variants}} \\
\cline{2-8}
\rule{0pt}{5ex}\singleheader{\textbf{Benchmark}} &
\shortstack[c]{\textbf{Nemotron-3.5}\\\textbf{Bolt-E4B imatrix}} &
\shortstack[c]{\textbf{Gemma4}\\\textbf{E4B-it}} &
\shortstack[c]{\textbf{Qwen3.5}\\\textbf{4B}} &
\shortstack[c]{\textbf{Nemotron-3}\\\textbf{Nano-4B}} &
\shortstack[c]{\textbf{Q4\_K\_M}\\\textbf{no imatrix}} &
\shortstack[c]{\textbf{IQ4\_XS}} &
\shortstack[c]{\textbf{IQ4\_NL}} \\
\midrule
\addlinespace[5pt]
\multicolumn{8}{@{}l}{\textbf{Conversational Proficiency}} \\
\midrule
MultiChallenge & \textbf{55.60} & 39.36 & 52.23 & 34.53 & 51.12 & 51.91 & 52.12 \\
RoleBench Custom & \textbf{71.21} & 60.51 & 52.35 & 10.86 & 65.21 & 69.07 & 69.86 \\
RoleMRC & \textbf{79.12} & 75.24 & 62.61 & 45.84 & 80.36 & 78.79 & 75.65 \\
EQ-Bench v3 & 38.15 & \textbf{39.77} & 32.07 & 19.45 & 37.42 & 38.19 & 38.28 \\
Brevity Benchmark & \textbf{82.84} & 78.49 & 75.17 & 70.98 & 81.29 & 81.30 & 82.04 \\
\addlinespace[5pt]
\multicolumn{8}{@{}l}{\textbf{Multi-Turn Tool Use}} \\
\midrule
BFCL-v3 & 67.31 & 65.03 & \textbf{70.56} & 59.07 & 64.32 & 65.95 & 67.68 \\
When2Call-MT & 60.50 & \textbf{61.12} & 42.88 & 35.96 & 58.06 & 60.67 & 60.92 \\
ToolSandbox & 74.13 & 68.31 & \textbf{74.51} & 59.78 & 70.23 & 70.44 & 69.88 \\
$\tau^2$-Bench & 52.32 & 38.72 & \textbf{61.74} & 14.89 & 46.95 & 42.05 & 48.74 \\
\addlinespace[5pt]
\multicolumn{8}{@{}l}{\textbf{Instruction Following \& Control}} \\
\midrule
IHEval & \textbf{82.40} & 66.61 & 62.87 & 53.75 & 83.50 & 83.19 & 83.65 \\
AgentIF & \textbf{22.95} & 20.05 & 22.24 & 21.25 & 24.33 & 25.32 & 25.46 \\
StructEval-T & \textbf{79.99} & 77.28 & 76.38 & 16.54 & 78.44 & 81.40 & 81.25 \\
IFBench & 64.00 & 42.67 & \textbf{66.67} & 58.58 & 59.79 & 59.00 & 63.33 \\
InverseIFEval & \textbf{45.26} & 39.75 & 22.04 & 36.02 & 42.69 & 44.47 & 43.08 \\
\addlinespace[5pt]
\multicolumn{8}{@{}l}{\textbf{Context Robustness}} \\
\midrule
RULER (64K) & 85.76 & 86.14 & \textbf{95.37} & 91.31 & 82.92 & 85.84 & 86.88 \\
LongMemEval & 69.92 & 67.95 & \textbf{76.60} & 27.15 & 64.30 & 70.20 & 72.80 \\
LongBench v2 (48K) & 48.67 & 45.70 & \textbf{58.28} & 37.42 & 44.37 & 47.33 & 47.67 \\
SysBench & 67.22 & 65.62 & \textbf{67.90} & 60.45 & 66.55 & 68.04 & 66.68 \\
\addlinespace[5pt]
\multicolumn{8}{@{}l}{\textbf{State \& Action Grounding}} \\
\midrule
GaRAGe & \textbf{40.73} & 30.91 & 31.78 & 24.80 & 38.85 & 38.74 & 39.95 \\
RAGTruth & \textbf{77.88} & 77.17 & 74.08 & 71.75 & 76.92 & 77.06 & 78.24 \\
RelyToolBench & \textbf{69.29} & 67.15 & 38.34 & 38.97 & 63.50 & 64.11 & 66.31 \\
\hline
\end{tabular}
\label{tab:q4-evals}
\end{table*}
\footnotetext{Gemma and Qwen 4-bit Checkpoints:
\href{https://huggingface.co/unsloth/gemma-4-E4B-it-GGUF/blob/main/gemma-4-E4B-it-Q4_K_M.gguf}{unsloth/gemma-4-E4B-it-GGUF}
and
\href{https://huggingface.co/unsloth/Qwen3.5-4B-GGUF/blob/main/Qwen3.5-4B-Q4_K_M.gguf}{unsloth/Qwen3.5-4B-GGUF}.}

%% file: sections/multilingual.tex
\section{Multilingual Evaluations}\label{sec:multilingual}
Since \ourmodel{} was post-trained using English only, we tested if the model retains the original model's multilingual capabilities. To test this, we evaluated \ourmodel on WMT24++ for 36 languages and MMLU-ProX on 12 languages. 

The results are shown in Tables \ref{tab:wmt24-xcomet} and \ref{tab:mmlu-prox-bf16} respectively. 
While \ourmodel retains the original model's machine translation quality, it shows slight but consistent drops on MMLU-ProX.
These results suggest that translation quality is preserved more consistently than multilingual knowledge and reasoning performance.

\begin{table*}[!htbp]
\centering
\definecolor{multiImproved}{HTML}{387353}
\definecolor{multiDegraded}{HTML}{963F4B}
\providecommand{\multidelta}[2]{}
\renewcommand{\multidelta}[2]{%
  \,{\fontsize{6}{7}\selectfont\textcolor{#1}{(#2)}}%
}

\begin{minipage}[t]{0.57\textwidth}
\vspace{0pt}
\centering
\caption{WMT24++ translation results for
Nemotron-3.5-Bolt-E4B BF16, measured by xCOMET and chrF
(both $\uparrow$). Both models use non-reasoning mode;
reasoning mode shows similar trends.
Parenthesized deltas are relative to Gemma4-E4B-it
for the same language, metric, and direction.
Green indicates improvement and red indicates degradation.}
\label{tab:wmt24-xcomet}

\fontsize{8}{9.8}\selectfont
\setlength{\tabcolsep}{2.5pt}
\renewcommand{\arraystretch}{1.08}
\begin{tabular}{@{}l|rr|rr@{}}
\toprule
 & \multicolumn{2}{c|}{\textbf{English $\rightarrow$ X}}
 & \multicolumn{2}{c}{\textbf{X $\rightarrow$ English}} \\
\cmidrule(lr){2-3}\cmidrule(l){4-5}
\textbf{Language}
 & \textbf{xCOMET} & \textbf{chrF}
 & \textbf{xCOMET} & \textbf{chrF} \\
\midrule
\texttt{ar\_EG}
 & 67.7\multidelta{multiDegraded}{-1.00}
 & 40.5\multidelta{multiImproved}{+0.20}
 & 75.5\multidelta{multiImproved}{+0.10}
 & 48.3\multidelta{multiDegraded}{-0.20} \\
\texttt{ar\_SA}
 & 73.9\multidelta{multiDegraded}{-0.70}
 & 40.7\multidelta{black}{+0.00}
 & 76.6\multidelta{black}{+0.00}
 & 49.2\multidelta{multiDegraded}{-0.10} \\
\texttt{bg\_BG}
 & 71.5\multidelta{black}{+0.00}
 & 57.4\multidelta{multiImproved}{+0.10}
 & 81.5\multidelta{multiDegraded}{-0.10}
 & 59.8\multidelta{multiDegraded}{-0.30} \\
\texttt{cs\_CZ}
 & 66.0\multidelta{multiImproved}{+1.60}
 & 49.8\multidelta{multiImproved}{+0.20}
 & 82.2\multidelta{multiImproved}{+0.10}
 & 59.1\multidelta{multiDegraded}{-0.20} \\
\texttt{da\_DK}
 & 85.6\multidelta{multiDegraded}{-0.20}
 & 63.5\multidelta{multiDegraded}{-0.20}
 & 89.3\multidelta{multiDegraded}{-0.10}
 & 64.1\multidelta{black}{+0.00} \\
\texttt{de\_DE}
 & 90.9\multidelta{multiImproved}{+0.20}
 & 58.6\multidelta{black}{+0.00}
 & 90.1\multidelta{black}{+0.00}
 & 58.8\multidelta{multiDegraded}{-0.10} \\
\texttt{el\_GR}
 & 67.0\multidelta{multiImproved}{+0.40}
 & 54.5\multidelta{black}{+0.00}
 & 81.0\multidelta{multiImproved}{+0.10}
 & 62.7\multidelta{black}{+0.00} \\
\texttt{es\_MX}
 & 84.6\multidelta{multiDegraded}{-0.20}
 & 66.6\multidelta{multiDegraded}{-0.50}
 & 89.3\multidelta{multiDegraded}{-0.10}
 & 66.4\multidelta{multiDegraded}{-0.20} \\
\texttt{et\_EE}
 & 41.4\multidelta{multiImproved}{+1.30}
 & 45.9\multidelta{multiImproved}{+0.30}
 & 68.3\multidelta{multiDegraded}{-0.40}
 & 56.3\multidelta{multiImproved}{+0.10} \\
\texttt{fi\_FI}
 & 75.0\multidelta{multiImproved}{+0.70}
 & 57.4\multidelta{black}{+0.00}
 & 81.6\multidelta{multiImproved}{+0.10}
 & 57.5\multidelta{multiDegraded}{-0.20} \\
\texttt{fr\_CA}
 & 79.4\multidelta{multiDegraded}{-0.70}
 & 67.5\multidelta{black}{+0.00}
 & 88.0\multidelta{multiDegraded}{-0.10}
 & 64.8\multidelta{multiDegraded}{-0.20} \\
\texttt{fr\_FR}
 & 79.5\multidelta{multiDegraded}{-0.30}
 & 63.2\multidelta{black}{+0.00}
 & 87.2\multidelta{multiDegraded}{-0.20}
 & 60.5\multidelta{multiDegraded}{-0.10} \\
\texttt{hi\_IN}
 & 67.2\multidelta{multiImproved}{+0.10}
 & 39.9\multidelta{black}{+0.00}
 & 83.6\multidelta{multiImproved}{+0.10}
 & 49.1\multidelta{multiDegraded}{-0.20} \\
\texttt{hr\_HR}
 & 74.2\multidelta{multiImproved}{+0.50}
 & 53.6\multidelta{multiImproved}{+0.30}
 & 83.1\multidelta{black}{+0.00}
 & 60.0\multidelta{multiDegraded}{-0.10} \\
\texttt{hu\_HU}
 & 61.4\multidelta{multiImproved}{+1.10}
 & 45.7\multidelta{multiImproved}{+0.10}
 & 78.5\multidelta{multiImproved}{+0.20}
 & 54.4\multidelta{multiDegraded}{-0.20} \\
\texttt{id\_ID}
 & 83.8\multidelta{black}{+0.00}
 & 63.2\multidelta{multiImproved}{+0.20}
 & 89.0\multidelta{multiDegraded}{-0.10}
 & 57.9\multidelta{multiImproved}{+0.10} \\
\texttt{it\_IT}
 & 82.1\multidelta{multiDegraded}{-0.80}
 & 64.5\multidelta{multiDegraded}{-0.30}
 & 88.6\multidelta{multiImproved}{+0.30}
 & 63.3\multidelta{multiDegraded}{-0.20} \\
\texttt{ja\_JP}
 & 82.1\multidelta{multiImproved}{+0.10}
 & 36.2\multidelta{multiImproved}{+0.80}
 & 84.5\multidelta{multiImproved}{+0.70}
 & 50.5\multidelta{multiDegraded}{-0.40} \\
\texttt{ko\_KR}
 & 84.3\multidelta{multiImproved}{+0.30}
 & 35.6\multidelta{multiDegraded}{-0.10}
 & 84.5\multidelta{multiImproved}{+0.30}
 & 54.1\multidelta{multiDegraded}{-0.10} \\
\texttt{lt\_LT}
 & 44.1\multidelta{multiImproved}{+0.80}
 & 41.0\multidelta{multiDegraded}{-0.20}
 & 68.4\multidelta{multiDegraded}{-0.30}
 & 47.6\multidelta{multiDegraded}{-0.10} \\
\texttt{lv\_LV}
 & 37.8\multidelta{multiImproved}{+1.60}
 & 43.5\multidelta{multiDegraded}{-0.50}
 & 66.6\multidelta{multiDegraded}{-0.20}
 & 52.4\multidelta{multiDegraded}{-0.20} \\
\texttt{nl\_NL}
 & 85.8\multidelta{multiImproved}{+0.20}
 & 58.9\multidelta{black}{+0.00}
 & 89.3\multidelta{multiImproved}{+0.10}
 & 60.3\multidelta{black}{+0.00} \\
\texttt{no\_NO}
 & 87.6\multidelta{multiImproved}{+0.20}
 & 67.0\multidelta{multiImproved}{+0.10}
 & 89.0\multidelta{multiDegraded}{-0.10}
 & 66.7\multidelta{multiDegraded}{-0.30} \\
\texttt{pl\_PL}
 & 78.3\multidelta{multiDegraded}{-0.10}
 & 48.5\multidelta{multiDegraded}{-0.20}
 & 84.9\multidelta{multiDegraded}{-0.40}
 & 56.9\multidelta{multiDegraded}{-0.20} \\
\texttt{pt\_BR}
 & 85.0\multidelta{multiDegraded}{-0.10}
 & 64.5\multidelta{multiDegraded}{-0.20}
 & 89.0\multidelta{black}{+0.00}
 & 62.5\multidelta{multiDegraded}{-0.10} \\
\texttt{pt\_PT}
 & 83.8\multidelta{multiDegraded}{-0.20}
 & 61.1\multidelta{black}{+0.00}
 & 88.2\multidelta{multiDegraded}{-0.40}
 & 60.9\multidelta{black}{+0.00} \\
\texttt{ro\_RO}
 & 73.3\multidelta{black}{+0.00}
 & 59.7\multidelta{black}{+0.00}
 & 82.2\multidelta{multiDegraded}{-0.10}
 & 63.6\multidelta{multiDegraded}{-0.10} \\
\texttt{ru\_RU}
 & 78.7\multidelta{multiImproved}{+0.20}
 & 51.5\multidelta{multiImproved}{+0.10}
 & 83.4\multidelta{black}{+0.00}
 & 53.0\multidelta{multiDegraded}{-0.10} \\
\texttt{sk\_SK}
 & 59.9\multidelta{multiImproved}{+1.40}
 & 45.4\multidelta{multiImproved}{+0.40}
 & 80.8\multidelta{multiImproved}{+0.20}
 & 56.0\multidelta{multiDegraded}{-0.10} \\
\texttt{sl\_SI}
 & 58.6\multidelta{multiImproved}{+1.70}
 & 48.2\multidelta{multiImproved}{+0.60}
 & 78.6\multidelta{multiImproved}{+0.10}
 & 58.0\multidelta{multiDegraded}{-0.20} \\
\texttt{sv\_SE}
 & 86.7\multidelta{multiImproved}{+0.60}
 & 63.2\multidelta{multiDegraded}{-0.10}
 & 89.4\multidelta{black}{+0.00}
 & 65.1\multidelta{black}{+0.00} \\
\texttt{th\_TH}
 & 81.4\multidelta{multiDegraded}{-0.60}
 & 51.2\multidelta{multiDegraded}{-0.40}
 & 86.3\multidelta{multiImproved}{+0.10}
 & 54.1\multidelta{multiDegraded}{-0.20} \\
\texttt{tr\_TR}
 & 74.8\multidelta{black}{+0.00}
 & 58.5\multidelta{multiImproved}{+0.10}
 & 81.8\multidelta{multiDegraded}{-0.40}
 & 57.9\multidelta{multiImproved}{+0.10} \\
\texttt{uk\_UA}
 & 76.4\multidelta{multiDegraded}{-0.30}
 & 54.2\multidelta{black}{+0.00}
 & 80.4\multidelta{black}{+0.00}
 & 58.3\multidelta{multiDegraded}{-0.20} \\
\texttt{vi\_VN}
 & 80.7\multidelta{multiDegraded}{-0.90}
 & 55.2\multidelta{multiDegraded}{-0.30}
 & 87.3\multidelta{multiImproved}{+0.40}
 & 55.6\multidelta{multiDegraded}{-0.30} \\
\texttt{zh\_CN}
 & 80.8\multidelta{multiDegraded}{-0.50}
 & 37.3\multidelta{multiImproved}{+0.10}
 & 87.7\multidelta{black}{+0.00}
 & 52.3\multidelta{multiDegraded}{-0.40} \\
\texttt{zh\_TW}
 & 80.3\multidelta{multiDegraded}{-0.80}
 & 31.0\multidelta{multiDegraded}{-0.10}
 & 89.2\multidelta{multiDegraded}{-0.10}
 & 56.4\multidelta{multiDegraded}{-0.20} \\
\midrule
Average
 & 74.4\multidelta{multiImproved}{+0.20}
 & 52.5\multidelta{black}{+0.00}
 & 83.4\multidelta{black}{+0.00}
 & 57.7\multidelta{multiDegraded}{-0.10} \\
\bottomrule
\end{tabular}
\end{minipage}
\hfill
\begin{minipage}[t]{0.40\textwidth}
\vspace{0pt}
\centering
\caption{MMLU-ProX results ($\uparrow$) for
Nemotron-3.5-Bolt-E4B BF16 with reasoning enabled
and disabled. Parenthesized deltas are relative to
Gemma4-E4B-it for the same language and mode.
Green indicates improvement and red indicates degradation.}
\label{tab:mmlu-prox-bf16}

\fontsize{8}{9.8}\selectfont
\setlength{\tabcolsep}{3pt}
\renewcommand{\arraystretch}{1.08}
\begin{tabular}{@{}l|r|r@{}}
\toprule
\textbf{Language}
 & \textbf{Reasoning}
 & \textbf{Non-reasoning} \\
\midrule
\texttt{ar}
 & 64.04\multidelta{multiDegraded}{-1.97}
 & 60.90\multidelta{multiDegraded}{-1.69} \\
\texttt{de}
 & 66.15\multidelta{multiDegraded}{-1.59}
 & 64.33\multidelta{multiDegraded}{-1.05} \\
\texttt{en}
 & 70.41\multidelta{multiDegraded}{-0.57}
 & 69.34\multidelta{multiDegraded}{-0.69} \\
\texttt{es}
 & 67.24\multidelta{multiDegraded}{-1.33}
 & 65.11\multidelta{multiDegraded}{-0.76} \\
\texttt{fr}
 & 66.99\multidelta{multiDegraded}{-1.30}
 & 65.01\multidelta{multiDegraded}{-1.08} \\
\texttt{hi}
 & 64.72\multidelta{multiDegraded}{-1.73}
 & 61.49\multidelta{multiDegraded}{-1.52} \\
\texttt{it}
 & 66.97\multidelta{multiDegraded}{-1.67}
 & 65.26\multidelta{multiDegraded}{-1.00} \\
\texttt{ja}
 & 64.49\multidelta{multiDegraded}{-1.87}
 & 62.17\multidelta{multiDegraded}{-1.64} \\
\texttt{ko}
 & 64.51\multidelta{multiDegraded}{-1.77}
 & 61.90\multidelta{multiDegraded}{-1.57} \\
\texttt{pt}
 & 66.59\multidelta{multiDegraded}{-1.49}
 & 64.70\multidelta{multiDegraded}{-1.56} \\
\texttt{vi}
 & 66.56\multidelta{multiDegraded}{-1.35}
 & 64.59\multidelta{multiDegraded}{-0.90} \\
\texttt{zh}
 & 65.17\multidelta{multiDegraded}{-1.61}
 & 63.32\multidelta{multiDegraded}{-1.12} \\
\midrule
Average
 & 66.15\multidelta{multiDegraded}{-1.52}
 & 64.01\multidelta{multiDegraded}{-1.22} \\
\bottomrule
\end{tabular}
\end{minipage}
\end{table*}

\FloatBarrier

%% file: sections/multimodal.tex
\section{Multimodal Evaluations}\label{sec:multimodal}
Although the model was trained text-only, we found that the model retains the original model's multimodal capabilities. 

\paragraph{Audio understanding and spoken interaction.}
We evaluate \ourmodel and Gemma4-E4B-it on two popular speech benchmarks: VoiceBench~\citep{chen2024voicebench} and AudioMC~\citep{gosai2025audiomultichallenge}. VoiceBench is a single-turn speech benchmark comprising subsets of benchmarks covering multiple-choice question answering (OpenBookQA, MMSU), open-ended question answering (CommonEval, SD-QA, AlpacaEval, WildVoice), reasoning (BBH), instruction following (IFEval), and safety (AdvBench). AudioMC is a multi-turn speech benchmark built on the text benchmark MultiChallenge, with an additional evaluation of voice editing that assesses a model's ability to track information as users provide updates during a conversation.

\paragraph{Vision capabilities.}
In contrast, \ourmodel exhibits slight but consistent degradation in vision capabilities. We attribute this degradation to misalignment between the vision encoder and the language model introduced by text-only training. However, the declines are not substantial on most vision benchmarks, suggesting that \ourmodel still retains visual understanding capabilities of the original model. We attempted additional training of the vision encoder alone but observed no meaningful improvements. Restoring and further improving vision capabilities remain part of future work.

\begin{table*}[!htbp]
\centering
\caption{Audio and vision benchmark results in reasoning and non-reasoning modes. Gemma4-E4B-it scores are measured in our evaluation setup\textsuperscript{a}. Bold indicates the best score among models within each mode. Arrows indicate whether higher ($\uparrow$) or lower ($\downarrow$) scores are better; CoVoST excludes Chinese. Parenthesized deltas are Bolt minus the measured Gemma4-E4B-it score in the same mode; Green indicates improved performance and red indicates degraded performance, accounting for whether higher or lower scores are better.}
\label{tab:audio-vision-results}
\definecolor{boltDeltaPositive}{HTML}{387353}
\definecolor{boltDeltaNegative}{HTML}{963F4B}
\newcommand{\boltdeltapos}[1]{\,{\fontsize{6}{7}\selectfont\textcolor{boltDeltaPositive}{(+#1)}}}
\newcommand{\boltdeltaposbad}[1]{\,{\fontsize{6}{7}\selectfont\textcolor{boltDeltaNegative}{(+#1)}}}
\newcommand{\boltdeltaneg}[1]{\,{\fontsize{6}{7}\selectfont\textcolor{boltDeltaNegative}{(-#1)}}}
\newcommand{\boltdeltaneggood}[1]{\,{\fontsize{6}{7}\selectfont\textcolor{boltDeltaPositive}{(-#1)}}}
\fontsize{8}{9.8}\selectfont
\setlength{\tabcolsep}{3pt}
\renewcommand{\arraystretch}{1.12}
\begin{tabularx}{\textwidth}{@{}>{\raggedright\arraybackslash}X|rrr|rrr@{}}
\toprule
 & \multicolumn{3}{c|}{\textbf{Reasoning}}
 & \multicolumn{3}{c}{\textbf{Non-reasoning}} \\
\cmidrule(lr){2-4}\cmidrule(l){5-7}
\textbf{Benchmark} &
\shortstack{\textbf{Nemotron-3.5}\\\textbf{Bolt-E4B}} &
\shortstack{\textbf{Gemma4}\\\textbf{E4B-it}} &
\shortstack{\textbf{Qwen3.5}\\\textbf{4B}} &
\shortstack{\textbf{Nemotron-3.5}\\\textbf{Bolt-E4B}} &
\shortstack{\textbf{Gemma4}\\\textbf{E4B-it}} &
\shortstack{\textbf{Qwen3.5}\\\textbf{4B}} \\
\midrule
\multicolumn{7}{@{}l}{\textbf{Audio}} \\
\midrule
Audio MultiChallenge $\uparrow$ & \textbf{25.22}\boltdeltapos{6.19} & 19.03 & --- & \textbf{19.25}\boltdeltapos{2.21} & 17.04 & --- \\
Voicebench $\uparrow$ & \textbf{80.30}\boltdeltapos{2.10} & 78.20 & --- & \textbf{72.60}\boltdeltapos{0.90} & 71.70 & --- \\
CoVoST (BLEU) $\uparrow$ & 36.14\boltdeltaneg{0.81} & \textbf{36.95} & --- & \textbf{37.61}\boltdeltapos{0.59} & 37.02 & --- \\
FLEUR (WER/CER) $\downarrow$ & 10.16\boltdeltaposbad{0.38} & \textbf{9.78} & --- & 10.43\boltdeltaposbad{0.26} & \textbf{10.17} & --- \\
OpenASR (WER) $\downarrow$ & --- & --- & --- & \textbf{9.28}\boltdeltaneggood{0.25} & 9.53 & --- \\
\addlinespace[5pt]
\multicolumn{7}{@{}l}{\textbf{Vision}} \\
\midrule
OmniDocBench 1.5 $\downarrow$ & 0.42\boltdeltaposbad{0.21} & 0.21 & \textbf{0.17} & 0.47\boltdeltaposbad{0.23} & 0.24 & \textbf{0.20} \\
MMMU-Pro Vision $\uparrow$ & 42.02\boltdeltaneg{5.61} & 47.63 & \textbf{66.13} & 39.08\boltdeltaneg{7.10} & 46.18 & \textbf{56.18} \\
MathVision $\uparrow$ & 46.25\boltdeltaneg{2.04} & \textbf{48.29} & --- & 43.72\boltdeltaneg{0.36} & 44.08 & \textbf{53.45} \\
MedXpertQA Vision $\uparrow$ & 26.85\boltdeltaneg{1.35} & 28.20 & \textbf{41.85} & 23.10\boltdeltaneg{0.65} & 23.75 & \textbf{35.35} \\
\bottomrule
\end{tabularx}
\par\smallskip
\begin{minipage}{\textwidth}
\footnotesize
\textsuperscript{a}We report our own measurements for Gemma4-E4B-it to ensure consistent evaluation across models; we observe discrepancies with the results reported in Google's model card (\texttt{huggingface.co/google/gemma-4-E4B-it}).
\end{minipage}
\end{table*}

The results in Table~\ref{tab:audio-vision-results} show that \ourmodel improves upon Gemma4-E4B-it on both VoiceBench and AudioMC, with reasoning enabled and disabled. Both models support speech inputs of up to 30 seconds. In AudioMC, 23.36\% of user turns exceed this limit, and 53.1\% of conversations contain at least one such turn. Our reported AudioMC results are therefore subject to this per-turn input limit and do not reflect evaluation on the complete, unrestricted conversations. For the SD-QA subset of VoiceBench, we report the average of the GPT score and the PANDA score. We compute PANDA using \texttt{qa\_metrics} v0.2.42, which showed better agreement with GPT-based judgments in our evaluation.

%% file: sections/speculative_decoding.tex
\section{Speculative Decoding}\label{sec:speculation}

\ourmodel{} is released with a DFlash~\citep{dflash2025} parallel drafter.
The drafter proposes all draft tokens in one forward pass, which the target verifies in parallel; verification preserves the output distribution of the target~\citep{leviathan2023fast,chen2023accelerating}.
We measure drafting quality by the acceptance length (AL), the number of tokens generated per verification step, including the one contributed by the target (at most 8 for a draft length of 7).

\paragraph{Algorithm selection.}
Recent extensions of DFlash add a lightweight correction head that improves the mutual consistency of the drafted tokens.
We compare DFlash with LiLiCorr~\citep{rusanovsky2026lilicorrlightweightlikelihoodcorrelation}, a state-of-the-art extension that jointly scores candidate tokens across all drafted positions.
For \ourmodel{}, LiLiCorr increases AL by about 5\%.
However, because each decoding step of such a small model is short, even a lightweight head adds non-negligible overhead, and the two drafters achieve comparable per-user decode speed at concurrency 1.
We therefore adopt DFlash, which minimizes the drafting overhead.

\paragraph{Drafter and training.}
Our DFlash drafter is a five-layer Transformer with 360M parameters (excluding the token embeddings and LM head, which are reused from the target) and drafts seven tokens per step.
Attention is bidirectional among the drafted positions and restricted to a sliding window of the most recent 1,024 positions.
The drafter is conditioned on hidden states from five intermediate layers of the target.
We train it in NVIDIA Model Optimizer (ModelOpt;~\citealp{nvidia-modelopt}) by distilling the next-token distribution of the target at every drafted position.
The training data comprise 3.79M conversations whose prompts are drawn from \href{https://huggingface.co/datasets/nvidia/Nemotron-Post-Training-Dataset-v2}{Nemotron-Post-Training-Dataset-v2} and whose responses are regenerated by the target at temperature 1.0, in both reasoning-enabled and reasoning-disabled modes.

\paragraph{Quantization.}
To speed up drafting on GPUs with lower memory bandwidth, typical of local deployments, we apply post-training NVFP4 quantization to the drafter using GPTQ~\citep{frantar2023gptq} in ModelOpt.
This reduces the weights read per drafting step, including the LM head, by about 3.6$\times$, with at most a 2\% reduction in AL.
By default, activations remain in BF16 (W4A16); an alternative variant of the checkpoint also quantizes them to NVFP4 (W4A4), which yields additional speedup at higher concurrency on GPUs with native NVFP4 support.
Table~\ref{tab:specdec-quant} reports the speedups of both variants on DGX Spark.

\begin{table}[h]
\centering
\caption{Weights read per drafting step (including the LM head), AL, and per-user decode speedup of the NVFP4 drafters relative to the BF16 drafter on DGX Spark. AL is measured at concurrency 1.}
\label{tab:specdec-quant}
\begin{tabular}{lccccc}
\hline
 & Weights & & \multicolumn{3}{c}{Speedup at concurrency} \\
\cline{4-6}
Drafter & (GB) & AL & 1 & 16 & 32 \\
\hline
BF16 & 2.06 & 2.99 & 1.00$\times$ & 1.00$\times$ & 1.00$\times$ \\
NVFP4 W4A16 (default) & 0.58 & 2.99 & 1.13$\times$ & 1.05$\times$ & 1.01$\times$ \\
NVFP4 W4A4 & 0.58 & 2.93 & 1.11$\times$ & 1.09$\times$ & 1.06$\times$ \\
\hline
\end{tabular}
\end{table}

\paragraph{Evaluation.}
Table~\ref{tab:specdec-al} reports AL of the released drafter on MT-Bench~\citep{zheng2023judging}, the qualitative split of SPEED-Bench~\citep{2026speedbench}, and SPEED-Bench prompts with 32K-token inputs.

\begin{table}[h]
\centering
\caption{AL of the released drafter (NVFP4, W4A16) with reasoning enabled, on one GB300 GPU at concurrency 16 with up to 4,096 output tokens. At \textit{temp}\,=\,1, the target samples while the drafter drafts greedily.}
\label{tab:specdec-al}
\begin{tabular}{lrr}
\hline
Benchmark & \textit{temp}\,=\,0 & \textit{temp}\,=\,1 \\
\hline
MT-Bench & 4.01 & 3.65 \\
SPEED-Bench, qualitative split & 3.88 & 3.50 \\
SPEED-Bench, 32K-token inputs & 3.42 & 3.00 \\
\hline
\end{tabular}
\end{table}

\FloatBarrier

%% file: sections/serving.tex
\section{Inference Serving}\label{sec:serving}

\subsection{NVIDIA NIM}\label{sec:serving-nim}

We package the BF16 checkpoint of \ourmodel{} as a downloadable NVIDIA NIM
microservice built on vLLM~0.30.0~\citep{kwon2023efficient}. The container image is published
for \texttt{amd64} and \texttt{arm64} and ships a single BF16 profile on one GPU
with the following defaults:
\begin{itemize}
  \item \textbf{Speculative decoding.} The DFlash~\citep{dflash2025} drafter (Section~\ref{sec:speculation}) is packaged next to the
        main weights and enabled by default with a draft length of 7, so no
        extra download or configuration is needed.
  \item \textbf{Reasoning and tool calling.} Reasoning is on by default. The
        \texttt{gemma4} reasoning parser returns the reasoning separately from
        the answer, and the \texttt{gemma4} tool-call parser enables tool calling.
  \item \textbf{Context and modalities.} The context length is 65{,}536 tokens.
        Text, image, and audio inputs are accepted.
  \item \textbf{APIs.} OpenAI-compatible Chat Completions, Completions, and
        Responses endpoints, and the Anthropic-compatible Messages endpoint.
\end{itemize}

Three request-level controls are exposed through the chat template and the
serving engine: \texttt{enable\_thinking} turns reasoning on or off per request,
\texttt{thinking\_token\_budget} caps the number of reasoning tokens, and
\texttt{conversational\_mode} requests brief, spoken-style replies for voice
assistants. We validated the NIM on DGX Spark (GB10), RTX PRO 6000 and RTX PRO
4500 Blackwell Server Edition, and H100 80GB.

\paragraph{Speculative decoding speedup.}
We measure the effect of the packaged DFlash drafter by comparing two NIM
builds that differ only in whether the model profile includes the drafter; all
other serving settings are equal.
We use SPEED-Bench~\citep{2026speedbench} prompts from 11 categories (16 prompts per category
on RTX PRO 6000 and 8 on DGX Spark), streaming Chat Completions requests with
greedy decoding and \texttt{max\_tokens} of 1{,}024, at concurrency 1 and 8.
Per-user decode speed is $1000/\text{ITL}$, where ITL is the mean inter-token
latency in milliseconds. Acceptance length (AL) is reported by vLLM's
speculative decoding metrics and includes the token produced by the target
model. Table~\ref{tab:nim-dflash} reports means over the 11 categories from a single
run per configuration.

\begin{table}[h]
\centering
\caption{DFlash speedup in the \ourmodel{} NIM on SPEED-Bench
(BF16, one GPU, greedy decoding, \texttt{max\_tokens}=1{,}024). Per-user decode
speed is in output tokens per second; request latency is the mean end-to-end
latency per request.}
\label{tab:nim-dflash}
\begin{tabular}{llcccccc}
\hline
 & & & \multicolumn{3}{c}{Per-user decode speed (tok/s)} & \multicolumn{2}{c}{Request latency (s)} \\
\cline{4-8}
Reasoning & GPU & Conc. & No drafter & DFlash & Speedup & No drafter & DFlash \\
\hline
On  & RTX PRO 6000 & 1 & 124.7 & 341.2 & 2.74$\times$ & 5.3  & 2.0 \\
On  & RTX PRO 6000 & 8 & 115.7 & 313.7 & 2.71$\times$ & 5.9  & 2.2 \\
On  & DGX Spark    & 1 &  20.1 &  71.6 & 3.56$\times$ & 31.7 & 9.1 \\
On  & DGX Spark    & 8 &  23.3 &  66.3 & 2.85$\times$ & 28.7 & 10.1 \\
\hline
Off & RTX PRO 6000 & 1 & 126.5 & 302.0 & 2.39$\times$ & 3.4  & 1.4 \\
Off & RTX PRO 6000 & 8 & 117.8 & 273.3 & 2.32$\times$ & 3.7  & 1.6 \\
Off & DGX Spark    & 1 &  20.1 &  62.5 & 3.11$\times$ & 20.4 & 6.4 \\
Off & DGX Spark    & 8 &  23.4 &  57.2 & 2.45$\times$ & 18.6 & 7.0 \\
\hline
\end{tabular}
\end{table}

With reasoning on, DFlash reaches an AL of 3.8 and speeds up per-user decoding
by 2.7$\times$ on RTX PRO 6000 and 3.6$\times$ on DGX Spark at concurrency 1.
The speedup holds at concurrency 8 (2.7$\times$ and 2.9$\times$). With
reasoning off, the AL drops to 3.4--3.6 and the speedup to 2.3--3.1$\times$.
Because speculative decoding shortens decoding rather than prefill, the gain in
end-to-end request latency follows the decode speedup, while time to first
token changes by less than 100\,ms in all settings.

\paragraph{Serving overhead.}
In a matched comparison on DGX Spark against a standalone vLLM 0.30.0 server
with the same model and drafter weights, chat template, and scheduler limits,
the NIM delivered the same request throughput: per-run NIM-to-vLLM ratios
ranged from 0.99 to 1.01 at concurrency 1 and 4.

\subsection{Quantized Local Inference}
\label{sec:quantized-local-inference}

We compare \ourmodel with DFlash against
Gemma-4-E4B-it with its MTP drafter\footnote{\url{https://huggingface.co/google/gemma-4-E4B-it-assistant}} on the coding subset of SPEED-Bench, using AIPerf with a batch size of 1.
Both target models use Q4\_K\_M quantization, and the DFlash drafter uses BF16 precision.
As shown in Table~\ref{tab:quantized-coding-throughput},
\ourmodelshort achieves 272.6 tokens/s on RTX 5070, compared with 225.9 tokens/s for Gemma with MTP, corresponding to $1.21\times$ the throughput.
On RTX Spark N1X, \ourmodelshort achieves 124.4 tokens/s vs. 118.4 tokens/s, corresponding to $1.05\times$ the throughput.

\begin{table}[t]
  \centering
  \caption{
    Generation throughput on the coding subset of
    SPEED-Bench, measured using AIPerf with a batch size of 1.
    Both target models use Q4\_K\_M quantization;
    Bolt uses a BF16 DFlash drafter.
    Relative throughput is computed separately for each
    hardware platform against Gemma-4-E4B-it with MTP.
  }
  \label{tab:quantized-coding-throughput}
  \small
  \begin{tabular}{lrr}
    \hline
    Model and decoder
      & RTX 5070 & RTX Spark N1X \\
      & (tokens/s) & (tokens/s) \\
    \hline
    Gemma-4-E4B-it + MTP
      & 225.9 & 118.4 \\
    \ourmodel + DFlash
      & \textbf{272.6} & \textbf{124.4} \\
    \hline
    Relative throughput
      & $\mathbf{1.21\times}$ & $\mathbf{1.05\times}$ \\
    \hline
  \end{tabular}
\end{table}

%% file: sections/conclusion.tex
\section{Conclusion}
We presented \ourmodel, an open small language model designed for fast conversational agents. Built on \parentmodel, we apply reinforcement learning to target multi-turn tool use, instruction following, and output formatting while retaining its conversational strengths. Training in both reasoning and non-reasoning modes, together with verbosity control, supports applications that need concise responses and reliable actions under tight latency constraints.

The release combines these training choices with practical deployment options.
With DFlash, \ourmodelshort achieves $2.74\times$ the
per-user decode speed of inference without speculative
decoding in BF16 NIM serving on RTX PRO 6000 with
reasoning enabled at concurrency~1.
On the coding subset of SPEED-Bench, Q4\_K\_M
\ourmodelshort with DFlash achieves $1.21\times$ the
generation throughput of Q4\_K\_M Gemma-4-E4B-it
with MTP on RTX 5070 at batch size~1.

%% file: sections/contributors.tex
\section*{Contributors}

We thank the following people for their invaluable contributions to \ourmodel.

Adi Margolin, Aditya Malte, Aishwarya Chhabra, Alex Hodson, Ameya Sunil Mahabaleshwarkar, Ambrish Dantrey, Amy Shen, Andrii Skliar, Annamalai Chockalingam, Anusha Pant, Ashwath Aithal, Ashwin Kumar, Besmira Nushi, Bhavesh Pawar, Bryan Catanzaro, Bryan Yu, Chantal Hwang, Chenhan Yu, Chris Alexiuk, Conner Currie, Cynthia Gao, David Mosallanezhad, Dina Yared, Ehsan Hosseini Asl, Eileen Long, Erick Galinkin, Felix Friedmann, Giedrius Burachas, Gorkem Batmaz, Grzegorz Chlebus, Guillermo Siman, Haifeng Qian, Hao Guo, Hayley Ross, Hrishith Thadicherla, Hui Wang, Huiying Li, Ike Nnoli, Irina Filatova, Iris Liu, Ji-Eun Lee, Jian Leong, Jian Zhang, Jiaqi Zeng, Joey Conway, Jonathan Cohen, Jose Vergara, Julien Veron Vialard, Kari Briski, Katherine Luna, Kedar Pandurang Potdar, Kevin Ning, Khalil Ben Khaled, Krzysztof Pawelec, Laura Dykes, Leili Tavabi, Lily Lee, Linda Liu, Liwei Ding, Maka Dong, Manoj Kilaru, Maria Bagdasarova, Matthew Lee, Meredith Price, Micah Schaffer, Michael Boone, Michael Fukuyama, Michael Hasper, Nancy Agarwal, Negar Habibi, Oleksii Kuchaiev, Oluwatobi Olabiyi, Pablo Ribalta, Payal Motwani, Pinky Xu, Prasoon Varshney, Pritam Biswas, Puhui Meng, Rameshwar Shivbhakta, Ran Zilberstein, Reka Kovacs, Richard Mazzarese, Rishabh Garg, Rita Fernandes Neves, Rochelle Woods, Ruby Ye, Ruvaid Virk, Ryan Hill, Saeid Motiian, Seph Mard, Shai Ben Nun, Shaona Ghosh, Sharad Yadav, Sudeep Sabnis, Suhas Hariharapura Sheshadri, Terry Kong, Thanh Do, Thinh Le, Udi Karpas, Viet Anh Trinh, Vinay Raman, Vinita Sharma, Warren Drones, Wenwen Gao, Whei Hsueh, Will Efstratis, Yoshi Suhara, Yunsheng Liu, Zhenming Wang, Zijia Chen, Zuhair Ahmed.

\newpage